\documentclass{ieeeaccess}

\usepackage{kotex}
\usepackage{verbatim}
\usepackage{multirow}
\usepackage{amsmath}
\usepackage{microtype}
\usepackage{graphicx}
\usepackage{subfigure}
\usepackage{booktabs} 
\usepackage[
  separate-uncertainty = true,
  multi-part-units = repeat
]{siunitx}

\usepackage{cite}
\usepackage{amsmath,amssymb,amsfonts}
\usepackage{algorithmic}
\usepackage{graphicx}
\usepackage{textcomp}
\def\BibTeX{{\rm B\kern-.05em{\sc i\kern-.025em b}\kern-.08em
    T\kern-.1667em\lower.7ex\hbox{E}\kern-.125emX}}
\begin{document}
\history{Received August 23, 2021, accepted October 25, 2021, date of publication November 4, 2021, date of current version November 9, 2021.}
\doi{10.1109/ACCESS.2021.3125335}

\title{Discovering Latent Representations of Relations for Interacting Systems}
\author{\uppercase{Dohae Lee}\authorrefmark{1},
\uppercase{Young Jin Oh\authorrefmark{2}, and In-Kwon Lee}.\authorrefmark{3}}
\address[1]{Yonsei University, Seoul (e-mail: dlehgo1414@gmail.com)}
\address[2]{LG Electronics, Seoul (e-mail: skrcjstk@gmail.com)}
\address[3]{Yonsei University, Seoul (e-mail: iklee@yonsei.ac.kr)}
\tfootnote{
This research was supported by the MSIT(Ministry of Science and ICT), Korea, under the ITRC(Information Technology Research Center) support program(IITP-2021-2018-0-01419) supervised by the IITP(Institute for Information and Communications Technology Planning and Evaluation) and the National Research Foundation of Korea(NRF) grant funded by the Korea government(MSIT). (No. NRF-2020R1A2C2014622)
}

\markboth
{D. Lee \headeretal: DisCovering Latent Representations of Relations for Interacting Systems}
{D. Lee \headeretal: DisCovering Latent Representations of Relations for Interacting Systems}

\corresp{Corresponding author: In-Kwon Lee (e-mail: iklee@yonsei.ac.kr).}

\begin{abstract}
Systems whose entities interact with each other are common. 
In many interacting systems, it is difficult to observe the relations between entities which is the key information for analyzing the system. 
In recent years, there has been increasing interest in discovering the relationships between entities using graph neural networks. 
However, existing approaches are difficult to apply if the number of relations is unknown or if the relations are complex. 
We propose the DiScovering Latent Relation (DSLR) model, which is flexibly applicable even if the number of relations is unknown or many types of relations exist.
The flexibility of our DSLR model comes from the design concept of our encoder that represents the relation between entities in a latent space rather than a discrete variable and a decoder that can handle many types of relations.
We performed the experiments on synthetic and real-world graph data with various relationships between entities, and compared the qualitative and quantitative results with other approaches. 
The experiments show that the proposed method is suitable for analyzing dynamic graphs with an unknown number of complex relations. 
\end{abstract}

\begin{keywords}
graph neural network, relational inference, unsupervised learning
\end{keywords}

\titlepgskip=-15pt

\maketitle

\section{Introduction}
\label{sec:introduction}
\PARstart{M}{ost} entities in nature are related to each other, interacting based on their relationship, and 
changing their states over time based on such interactions.
An interacting system can be represented as a dynamic graph \cite{dynamicgraph} in which the attributes of the entities change over time. 

In most cases, although changes in the states of the entities can be observed, the relations among the entities have rarely been observed.
It is important to discover these hidden relationships between entities in a dynamic graph, which helps us to better understand the system and predict future states.
For example, inferring the relationship between physically interacting objects helps us understand the entire system and predict the future movements of objects. 
We can also analyze the specific motions by inferring the intrinsic relationships between the joints of the human body, and more accurately predict the future motion of each joint.

\begin{figure}[t]
\vskip 0.0in
\begin{center}
\centerline{\includegraphics[width=\columnwidth]{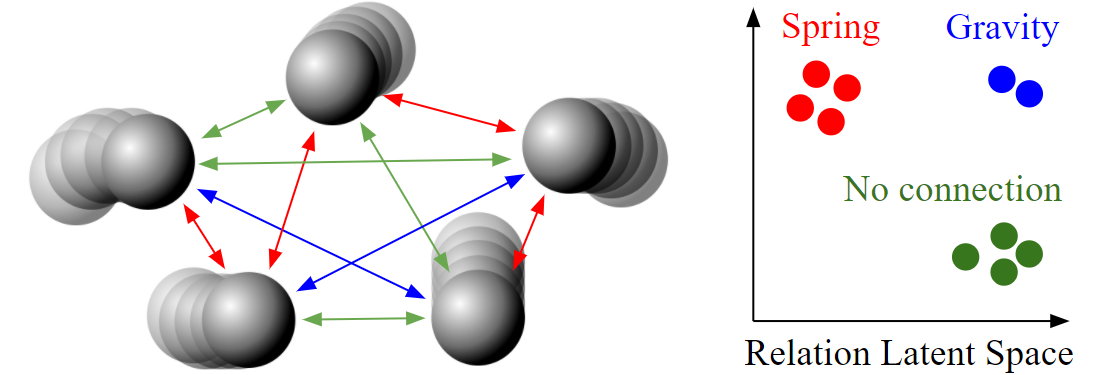}}
\caption{Left: Graph representing the dynamics of physical objects with different relations between objects. Right: Different relations are clustered within the relation latent space of the DSLR.}
\label{Figure:teaser}
\end{center}
\vskip -0.2in
\end{figure}

Several attempts have recently been made to infer the relationships between entities by observing the states of such entities in a dynamic graph, and to predict the future states of entities based on their relationships \cite{NRI, fNRI, SUGAR, dNRI}. 
It is possible to cluster the relations between entities and accurately predict their future states.
However, the existing models are designed to have a dedicated update function for each relation, making it difficult to model the system if the number of relationships is not known in advance. 
In addition, existing models can only discretely classify different relations, making them difficult to apply to data with complex relations. Furthermore, although existing models can distinguish different relations, it is difficult to analyze the relationships between these relations, such as their similarities. 
These issues are not trivial because entities in nature may have a complex and unknown number of relations. 

To overcome these challenges, we propose a DiScovering Latent Relation (DSLR) model composed of two graph neural networks (GNNs) \cite{gnn1,gnn2,gnn3,gnn4}. With DSLR, a relation encoder, i.e., the first GNN, embeds the relations between entities in the latent space. The relation decoder, i.e., the second GNN, predicts the future states of the entities using the recognized relations. 
The concept of DSLR is depicted in Fig. \ref{Figure:teaser}.
The figure on the left shows the dynamics of physical objects interacting with each other. 
The color of the arrows between objects indicates the type of relation between objects.
Each relation between objects can be represented in the relation latent space shown on the right side of the figure, in which the same relations are close to each other, and different relations are distant from each other.

Unlike previous methods that cluster relations discretely using softmax, our DLSR can better explain complex relations by embedding relations into a relation latent space. 
Even if the number of relations is unknown in advance, DSLR can not only accurately discover the relations, it can also infer the number of relationships in an interacting system. 
In addition, the relation decoder of DSLR predicts the future states of entities for all relation types; thus, even if the number of relations is large or complex, there is no need to modify or scale the model up. 

The main contributions in our work consist of:
\begin{itemize}
    \item We design a new relational inference model that can be applied even when the number of relationships in the system is unknown.
    \item Our method is more efficient for the interacting system with very complex relationships between entities, such as real-world motion capture data and basketball data.  
\end{itemize}

The remainder of the paper is organized as follows. The related works for better understanding the proposed framework are discussed in Section \ref{sec:related_work}. Section \ref{sec:method} discusses the proposed DSLR method, and Section \ref{sec:experiments} shows the experimental results. Finally, Section \ref{sec:conclusion} offers the conclusion.

\section{Related Work}
\label{sec:related_work}
\subsection{Relational Inductive Bias}
With the advent of deep learning, many important problems have been solved in various fields, including image processing, natural language processing, and control \cite{imageDNN, langDNN, controlDNN, accesscnn}. 
In recent years, there has been increasing interest in using relational inductive biases with deep learning architectures to solve relational reasoning problems \cite{relationalInductiveBias}.
For example, to allow multiple agents to cooperate in maximizing common rewards, a CommNet \cite{commNet} model designed to allow agents to communicate with each other was proposed.
An interaction network \cite{IN} model was also proposed to predict the future movement of physically interacting objects. 
Some researchers have addressed a visual question answering task that requires consideration of the relationship between objects \cite{relationalReasoning}, and attempts have been made to solve various relation-based tasks using a self-attention technique \cite{vain, graphAttention, access_self_attention}. 
In addition, a few-shot learning technique that considers relational information has been proposed \cite{fewshotGNN}, and a method for distinguishing and grouping objects from images was proposed \cite{rNEM}.
In addition, a number of GNN-based methods \cite{VIN, discovering, fluid, visualGrounding, accessgnn} with a strong relational inductive bias have been proposed to solve various relational reasoning tasks. 
However, these methods did not explicitly infer the relations between entities.

\subsection{Inferring Explicit Relation}
Meanwhile, there have been attempts to explicitly infer the relations between entities. 
For example, several studies have explored how to infer relations in the fields of causal reasoning \cite{otherRelation1} and computational neuroscience \cite{otherRelation2, otherRelation3}. 
Methods for clustering the relations between entities and predicting the future states of such entities using a GNN \cite{gnn1, gnn2, gnn3, gnn4} with a dynamic graph showing the changes in the states of the entities over time have recently been reported. 
The neural relational inference (NRI) \cite{NRI} model, which is an unsupervised neural network model that can discretely distinguish relations between entities and predict the dynamics of such entities in interacting systems, has been proposed. 
The factorized NRI (fNRI) \cite{fNRI} model, which complements the NRI model and efficiently handles combinations of independent interactions, was also proposed, as was the SUGAR \cite{SUGAR} model, which modifies the NRI model to consider global interactions with various structured priors. 
In addition, a dynamic NRI (dNRI) model \cite{dNRI}, a method that can better infer the changing relations between entities in an interaction system, was developed.

These NRI-based methods can discover relations between entities and predict the future states of such entities in various interacting systems. 
However, because these methods are designed to handle each relation type with a dedicated update function, they may be difficult to apply to systems with unknown or large numbers of relationships. 
By contrast, our DSLR model is designed to represent relations in a latent space and handle all relations with a single update function; thus, it can be applied to systems with complex or unknown numbers of relations in a flexible manner.

\section{DiScovering Latent Relation Model}
\label{sec:method}

\begin{figure*}[ht]
\vskip 0.0in
\begin{center}
\centerline{\includegraphics[width=1.8\columnwidth]{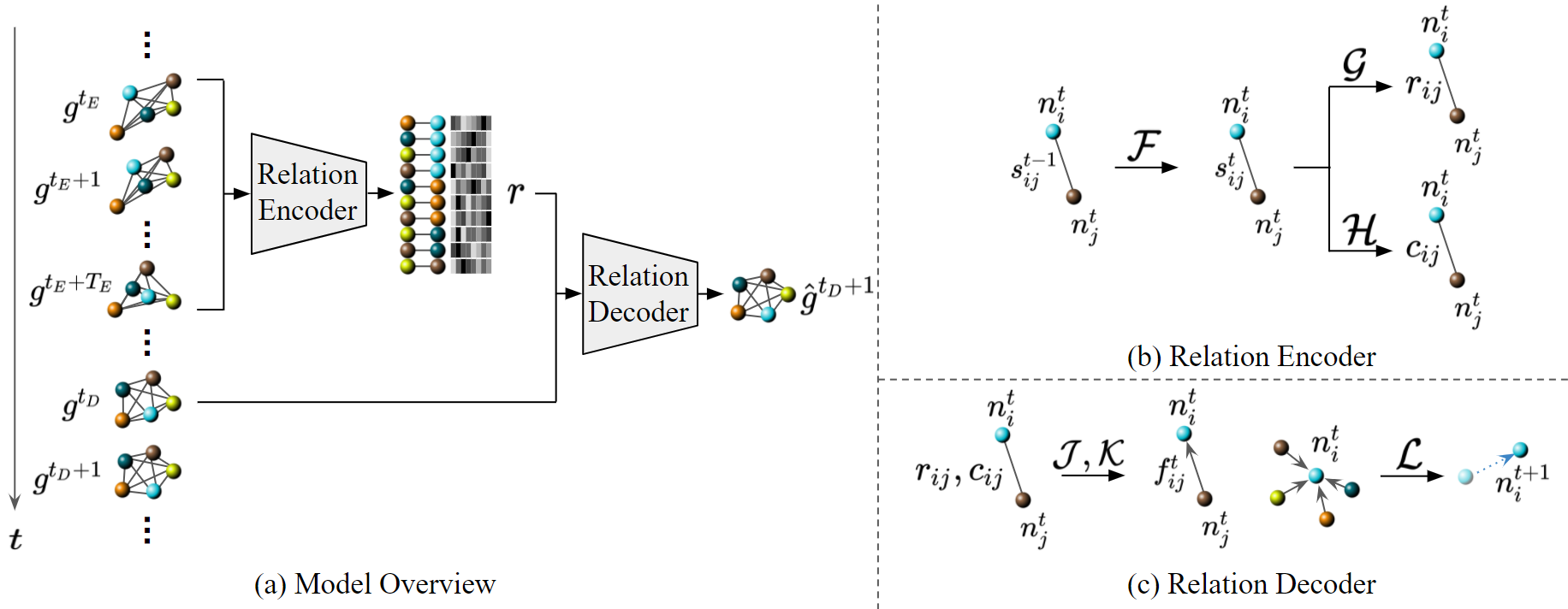}}
\caption{(a) Spherical objects representing entities, and lines connecting the objects representing their edges.
(b) Details of the relation encoder. (c) Details of the relation decoder.}
\label{Figure:model_overview}
\end{center}
\vskip -0.2in
\end{figure*}

\subsection{Model details}

The DSLR model consists of two subnetworks, i.e., a relation encoder and a relation decoder (see Fig. \ref{Figure:model_overview})

\subsubsection{Relation Encoder}
The first component of the DSLR is the relation encoder (see Fig. \ref{Figure:model_overview} (b)), which is a network that infers the relations between nodes by observing the states of the nodes (entities) in interacting systems for a certain period of time. 
The relation encoder first infers the edge states before inferring the relation states containing the relation information between two nodes. 
The edge state is information that serves as a clue to infer the relation state, which contains state information during several time steps of the edges between nodes. 
In a system with $N$ nodes, the edge state $s_{ij}^t$ between the $i$-th node $n_i^t$ and the $j$-th node $n_j^t$ at time $t$ is defined as follows:
\begin{equation}
    s_{ij}^{t} = \mathcal{F}\left ( n_i^t, n_j^t, s_{ij}^{t-1} \right ), \ \ \ i,j = 1, ..., N, \ \ \ i \neq j,
\label{edge_state}
\end{equation}
where $\mathcal{F}$ is a function that updates the edge state $s_{ij}^{t}$ to $s_{ij}^{t+1}$ from the edge state and node states of the previous time step.

Next, the relation encoder infers the relation state between the nodes based on the edge state. 
We use a reparameterization trick of a variational auto-encoder \cite{VAE} to enable the stochastic representation of the latent vectors. 
We model each component in the latent vector to have a prior with a normal distribution having a mean of zero and a standard deviation of 1.
When the graph is observed for $T_E$ time steps from the first time-step, the relation state $r_{ij}$ between the $i$-th and $j$-th nodes is obtained as follows:
\begin{equation}
    r_{ij} \sim \mathcal{G}(s_{ij}^{T_E}),
    \label{relation_state}
\end{equation}
where $\mathcal{G}$ is a function that models the distribution of the relation state from the edge state, and $\sim$ is a sampling operation using the reparameterization trick.

The relation encoder can infer the relation centrality, which is the importance of the relation. 
The relation encoder obtains the relation centrality from the edge state in the same way in which the relation state is inferred. 
The relation centrality $c_{ij}$ between nodes $i$ and $j$ is obtained as follows:
\begin{equation}
    c_{ij} = \sigma(\mathcal{H}(s_{ij}^{T_E})),
    \label{relation_centrality}
\end{equation}
where $\mathcal{H}$ is a function that infers the relation centrality from the edge state, and $\sigma$ is a sigmoid function. 
The relation centrality $c_{ij}$ is a scalar value between zero and 1, and a large $c_{ij}$ indicates that the relation between $i$ and $j$-th nodes is important, and vice versa. 

\subsubsection{Relation Decoder}
The relation decoder predicts the future states of the nodes from the current states of the nodes and the relation state between them, as inferred by the relation encoder. 
The relation decoder first infers the influences exchanged between nodes, aggregates all influences, and predicts the future state of the nodes when considering all effects applied to each node. 
The influence of the $j$-th node on the $i$-th node at time $t$, $f_{ij}^t$, is modeled as follows:
\begin{equation}
    f_{ij}^t = \mathcal{J}(c_{ij})\mathcal{K}(n_i^t, n_j^t, r_{ij}),
    \label{influence}
\end{equation}
where $\mathcal{K}$ is a function that computes the influence of nodes on each other from the states of the nodes and their relation state, and $\mathcal{J}$ is a function that calculates the noise multiplied by the influence according to the relation centrality $c_{ij}$. 
The noise applied to the influence $f_{ij}^t$ is obtained as follows:
\begin{equation}
    \mathcal{J}(c_{ij}) = 1+(1-c_{ij})\epsilon,
    \label{J}
\end{equation}
where $\epsilon$ is a random variable. 
That is, the larger the relation centrality (closer to 1) is, the less noise applied to the influence, and the smaller the relation centrality (closer to 0) is, the greater the amount of noise. 

Next, the future state of the $i$-th node is calculated as follows:
\begin{equation}
    n_i^{t+1} = n_i^t + \mathcal{L} \left ( n_i^t, \sum^N_{j, \ i \neq j} f_{ij}^t \right ). 
    \label{L}
\end{equation}
where $\mathcal{L}$ represents a function that predicts change in the node states.

\subsection{Relation Reasoning}

To cluster the relations between nodes when the number of relations is small, unsupervised clustering \cite{clustering} is conducted on the relation states embedded in the relation latent space. We train the $k$-means clustering model with the relation states obtained from the training data, and cluster the relation states of the test data with the trained model. 
If the number of relations is unknown, the number of relations can be inferred using the silhouette method \cite{silhouette}. The larger the silhouette score, the higher the probability that $k$ will be optimal, which means that $k$ with the highest silhouette score is the number of relations. 

\subsection{Random Sampling Trick}
\label{subsec:random-sampling-trick}
It is desirable that the relation state contains only information about the relationship between nodes. 
However, when training a model, if the input of the relation encoder and input of the relation decoder are not independent of each other, the relation states learn the compressed information of input trajectories of nodes instead of relational information, which is undesirable.
In a previous study \cite{fNRI}, a model in which the module inferring the relation learns the compressed version of the input trajectories rather than the relation is called a `compression model.' 

To prevent our relation encoder from becoming a compression model, we propose a random sampling trick, i.e., while training the model, the times required to infer the relation state and to compute the influence between nodes are randomly sampled. 
In Fig. \ref{Figure:model_overview} (a), the relation encoder infers the relation state between nodes by observing the graph from time $t_E$ for $T_E$ time-steps, and the relation decoder predicts the graph of the next $g^{t_D+1}$ from the inferred relation state $r$ and the graph $g^{t_D}$ at time $t_D$. 
A random sampling trick is a technique that randomly samples $t_E$ and $t_D$ each time during training.

\subsection{Training}
The DSLR model is trained without supervision of the relation states in an end-to-end manner, which is optimized based on four loss functions, i.e., the node prediction loss, KL divergence loss, relation standard deviation loss, and relation centrality loss. 

The node prediction loss $L_{NP}$ is the mean squared error between the predicted future states of the nodes from time $t_D$ for $T_D$ time steps and the true future states:
\begin{equation}
    L_{NP} = \sum_{t=t_D}^{t_D+T_D}\left(\hat{n}^t - n^t\right)^2,
    \label{node_prediction_loss}
\end{equation}
where $\hat{n}^t$ is the ground truth state of the nodes at time $t$, and ${n}^t$ is the state of the nodes at time $t$ predicted by the model. 

KL divergence loss induces the distribution of each component in the relation states to follow a normal distribution. 
KL divergence loss $L_{KL}$ is defined as follows:
\begin{equation}
    L_{KL} = -D_{KL}\left(P(r) \vert\vert N(0,1) \right),
    \label{KL_divergence_loss}
\end{equation}
where $P(r)$ represents the distribution of each component in the relation states, which are the outputs of the relation encoder, $N(0,1)$ represents the standard normal distribution, and $D_{KL}$ represents the Kullback–Leibler divergence. 

Our model deals with a dynamic graph with a static relation where the states of the nodes change over time, but the relation between nodes does not change. 
In a single dynamic graph, the relation between nodes is constant. 
Therefore, in the same graph, the relation between nodes is constant regardless of the initial time $t_E$ of the graph input to the relation encoder (see Fig. \ref{Figure:model_overview} (a)). 
The relation standard deviation loss $L_{SD}$ is the standard deviation of the relation states inferred by the relation encoder for a sequence of graphs randomly sampled $m$ times within a dynamic graph:
\begin{equation}
    L_{SD} = STD(r_1, r_2, ..., r_m),
    \label{standard_deviation_loss}
\end{equation}
where $STD$ indicates the standard deviation, and $r_i$ indicates the relation state recognized in the $i$-th sampled sequence of graphs. 

In (4-5), the larger the relation centrality $c_{ij}$, the smaller the noise $J(c_{ij})$ that occurs in the influences $f_{ij}$ between nodes. 
Conversely, as the relation centrality decreases, the noise of the influences exchanged between nodes increases. 
Therefore, for the model to correctly predict the future states of the nodes, it is trained in the direction of increasing the relation centrality to decrease $L_{NP}$.  Contrary to the tendency of the relation centrality to become larger, we add a relation centrality loss to the objective function, which leads th a decrease in the relation centrality:
\begin{equation}
    L_c = -log(1-c),
    \label{relation_centrality_loss}
\end{equation}
where $c$ denotes the relation centrality. While the node prediction loss $L_{NP}$ is designed to model the interacting system correctly, the relation centrality loss acts in the opposite manner. 
When the magnitude of the influence exchanged between nodes $f_{ij}$ is large, the magnitude of $L_{NP}$ is larger than that of $L_c$, and thus the relation centrality increases to reduce the noise of $f_{ij}$. 
Conversely, when $f_{ij}$ is small, the magnitude of $L_c$ becomes relatively larger than that of $L_{NP}$, and thus the relation centrality decreases. 
As a result, the relation with a large relation centrality has a large overall effect on the interacting system, whereas the relation with a small relation centrality has a small effect. 

We combined all of the proposed losses and set the final objective function $L$ as follows:
\begin{equation}
    L = \lambda_{NP}L_{NP} + \lambda_{KL}L_{KL} + \lambda_{SD} L_{SD} + \lambda_{c} L_{c},
    \label{total_loss}
\end{equation}
where the weights $\lambda_{NP}$, $\lambda_{KL}$, $\lambda_{SD}$, and $\lambda_{c}$ used for scaling each loss function were set to $\lambda_{NP}=1$, $\lambda_{KL}=0.1$, $\lambda_{SD}=1$, and $\lambda_{c}=0.001$. 

\subsection{Sparsity Prior}

In this section, we introduce a method to set the sparsity prior to DSLR. 
A graph in which most of the nodes are not connected to each other is called a sparse graph. 
Most interacting systems in nature are sparse, and thus it would be useful if the model can reflect the sparsity of the graph as a prior. 
To add a sparsity prior to the model, we redefine the relation centrality loss defined in \ref{relation_centrality_loss}. 
Letting $p$ be the sparsity prior of the graph, the relation centrality loss $L_c$ is then redefined as follows:
\begin{equation}
    L_c = -\delta_{c,p}\log\left(1-c\right) - (1-\delta_{c,p})\log\left(c\right).
    \label{sparsity_prior}
\end{equation}
If the relation centrality $c$ falls within a small $p$\% in the training batch, $\delta_{c,p} = 1$; otherwise, $\delta_{c,p}=0$. 
That is, the modified relation centrality loss induces a smaller relation centrality within a small $p$\% (which indicates no connection relation), and a larger relation centrality within a large $(1-p)$\% (which indicates a meaningful relation).

\section{Experiments}
\label{sec:experiments}
The DSLR was optimized using the Adam optimizer \cite{adam} and implemented using Pytorch \cite{pytorch}. 
The learning rate was scheduled using a one-cycle learning rate policy \cite{onecyclelr}. 
All experiments were conducted using a 3090 GPU, and the numerical results were the average of three trials with different seeds.

\begin{table*}[t]
\caption{Accuracy of estimating the types of relation between objects in physically simulated data (in \%).}
\label{table:relation_acc}
\vskip 0.1in
\begin{center}
\begin{small}
\begin{sc}
\begin{tabular}{lcccr}
\toprule
\multirow{2}{*}{Dataset} & \multicolumn{2}{c}{NRI} & \multirow{2}{*}{DSLR} \\
 & $\bar K = K$ & $\bar K = 2K$ \\
\midrule
Spring \& None              &92.52\scriptsize{$\pm$10.69} 
&87.49\scriptsize{$\pm$8.39}
&\textbf{95.33\scriptsize{$\pm$0.46}}  \\
Spring \& Gravity             &75.84\scriptsize{$\pm$11.71}
&\textbf{96.12\scriptsize{$\pm$1.62}}
&95.22\scriptsize{$\pm$0.30} \\
Gravity \& None               &\textbf{92.28\scriptsize{$\pm$1.28}}
&80.50\scriptsize{$\pm$6.87}
&90.55\scriptsize{$\pm$1.00}  \\
Spring \& Gravity \& None     &84.90\scriptsize{$\pm$0.50} 
&73.63\scriptsize{$\pm$2.10}
&\textbf{88.80\scriptsize{$\pm$3.10}}  \\
3 Spring \& None   &66.38\scriptsize{$\pm$6.72}
&64.94\scriptsize{$\pm$5.61}
&\textbf{84.35\scriptsize{$\pm$2.24}}   \\
3 Gravity \& None   &64.33\scriptsize{$\pm$4.02}
&66.43\scriptsize{$\pm$0.92}
&\textbf{74.10\scriptsize{$\pm$1.41}}   \\
\bottomrule
\end{tabular}
\end{sc}
\end{small}
\end{center}
\vskip -0.1in
\end{table*}

\subsection{Physics Simulation}

To evaluate our DSLR model, we conducted experiments using physically simulated data in which objects have various relationships with each other. 
In our experiment, we chose relationships types that are frequently used in previous studies \cite{ IN, VIN, NRI, paig, discovering}: \textit{spring}, \textit{gravity}, and \textit{none}. 
\textit{Spring} is a relation in which a spring force acts between objects: If the objects are far from each other, an attractive force acts on the objects; however if they are close, a repulsive force acts on them. 
\textit{Gravity} is a relation in which an attractive force acts according to the distance between objects; however we set the attractive force inversely proportional to the distance instead of the squared distance. 
\textit{None} indicates that there is no connection between objects where the nodes do not transmit any forces to each other. 
We generated physically simulated data with eight combinations of relation types for the experiments: (a) \textit{spring \& none}, (b) \textit{spring \& gravity}, (c) \textit{gravity \& none}, (d) \textit{spring \& gravity \& none}, (e) \textit{3 spring \& none}, (f) \textit{3 gravity \& none}, (g) \textit{100 spring}, and (h) \textit{100 spring \& 100 gravity}.
In (a)--(d), objects are connected by one of the corresponding relations. 
In (e) and (f), the objects are connected in one of the four relations based on the weak, moderate, and strong \textit{spring} or \textit{gravity} forces and \textit{none}. 
In (g), a system is composed of $100$ types of relations between objects, where the objects are connected by \textit{springs} with $100$ different coefficients. 
In (h), a system consists of $200$ types of \textit{spring} relations with 100 different coefficients and \textit{gravity} with $100$ different coefficients between nodes. 
Simulations were generated by assigning the potential between two objects as in \cite{discovering}. 
We conducted experiments to infer the relation between objects and predict the future trajectory of the objects in an interacting system in which a number of objects move in a complex manner. 
All datasets consisted of 5000 training sets, 500 validation sets, and 500 test sets. Each data consists of 99 time steps, of which 49 are observed to infer the relation. 
We set $m$ in \eqref{standard_deviation_loss} to 5 in all experiments using physically simulated data. 

\begin{table}[t]
\caption{Silhouette scores of DSLR by $K$ in physically simulated data.}
\label{table:relation_num}
\vskip 0.1in
\begin{center}
\begin{small}
\begin{sc}
\begin{tabular}{lcccccr}
\toprule
Data set & $K$=2 & $K$=3 & $K$=4 & $K$=5 \\
\midrule
Spring \& None              &\textbf{0.96} &0.93 &0.89 &0.73    \\
Spring \& Gravity             &\textbf{0.97} &0.96 &0.94 &0.91    \\
Gravity \& None               & \textbf{0.90} & 0.85 & 0.79 & 0.72 \\
Spring \& Gravity \& None     &0.72 &\textbf{0.83} &0.81 &0.78    \\
3 Spring \& None   & 0.62 & 0.64 & \textbf{0.73} & 0.70  \\
3 Gravity \& None   & 0.62 & 0.60 & \textbf{0.72} & 0.69  \\
\bottomrule
\end{tabular}
\end{sc}
\end{small}
\end{center}
\vskip -0.1in
\end{table}

We conducted comparative experiments using the NRI model \cite{NRI}, which can discover the relation between objects.
The NRI model was trained using the training schema proposed by fNRI \cite{fNRI} to prevent it from becoming a compression model, during which, the first half of the physical data sequence was used to infer the relation, and the second half was used for a future trajectory prediction. 
Because the NRI model requires a dedicated decoder for each relation, the number of relations must be entered into the model in advance. 
When comparing the accuracy of the relation inference, we assumed that the true number of relations $K$ is known to the NRI. 
We also compared the result if the wrong number of relations ($\bar K = 2 K$) is set to the NRI. 
When comparing trajectory prediction errors, both the NRI with the correct number of relations ($\bar K = K$) and the NRI with an incorrect number of relations ($\bar K = \lceil K/2 \rceil$, $\bar K = 2K$) were used in the experiment. 
Neither the NRI model nor our DSLR model were given the sparsity prior. 
The DSLR model was trained with $\epsilon$ in \eqref{influence} set to zero because we did not aim to infer the relation centrality in this experiment, except for the relation centrality verification experiments. 
The models were trained for $1000$ epochs. 

\begin{figure}[t]
\vskip 0.0in
\begin{center}
\centerline{\includegraphics[width=\columnwidth]{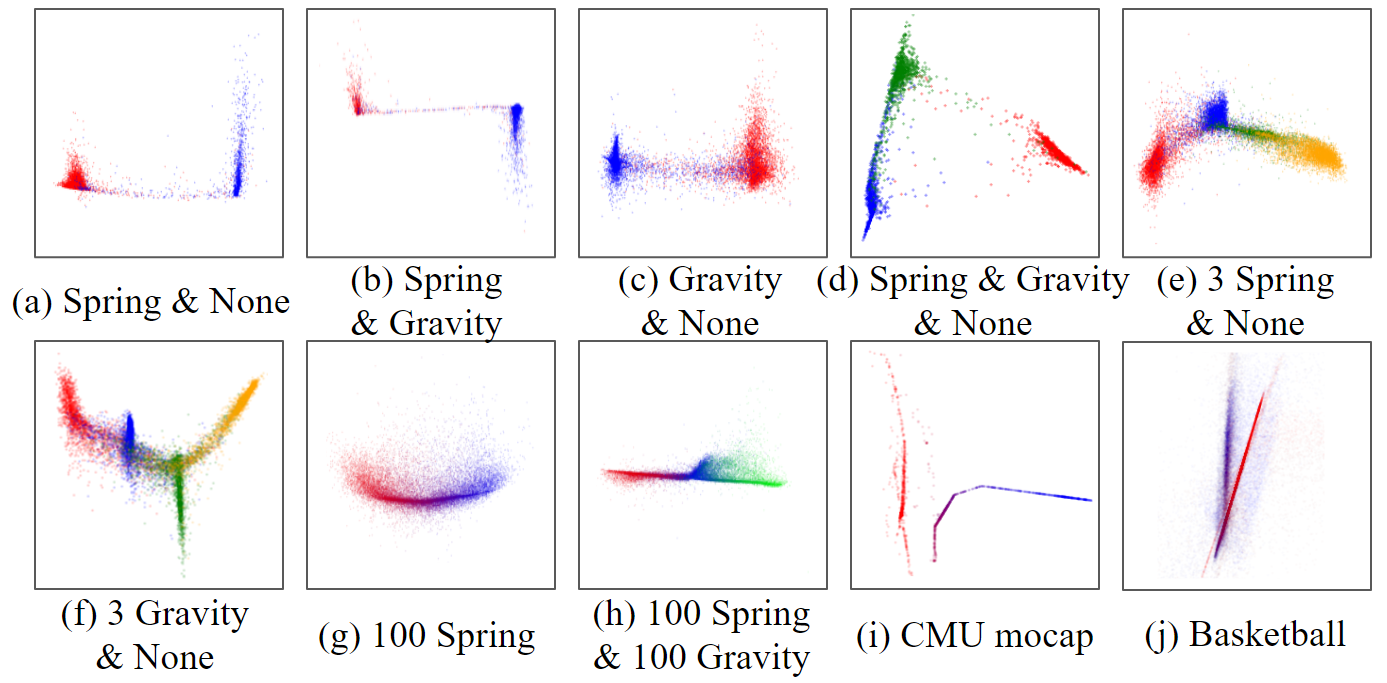}}
\caption{Latent representations of relation states reduced to a two-dimensional plane. (a) Red, \textit{spring}; blue, \textit{none}. (b) Red, \textit{gravity}; blue, \textit{spring}. (c) Red, \textit{gravity}; blue, \textit{none}. (d) Red, \textit{spring}; blue, \textit{gravity}; and green, \textit{none}. (e) Red, \textit{none}; blue, \textit{weak spring}; green, \textit{moderate spring}; and yellow, \textit{strong spring}. 
(f) Red, \textit{none}; blue, \textit{weak gravity}; green, \textit{moderate gravity}; and yellow, \textit{strong gravity}. 
(g) The stronger the \textit{spring} is, the bluer it is, and the weaker the \textit{spring} is , the redder it is.
(h) The stronger the \textit{spring} is, the redder it is; the stronger the \textit{gravity} is, the greener it is; and the weaker the force is, the bluer it is.
(i)--(j) The higher the relation centrality is, the redder it is; the lower relation centrality is, the bluer it is. 
}
\label{Figure:relation_latent_space}
\end{center}
\vskip -0.3in
\end{figure}

\begin{table*}[t]
\caption{Error in forecasting of future trajectories in physically simulated data.}
\label{table:trajectory_error}
\vskip 0.1in
\begin{center}
\begin{small}
\begin{sc}
\begin{tabular}{lccccr}
\toprule
\multirow{2}{*}{Dataset} & \multicolumn{3}{c}{NRI} & \multirow{2}{*}{DSLR} \\
 &$\bar K=\lceil K/2 \rceil$ & $\bar K=K$ & $\bar K=2K$ \\
\midrule
Spring \& None             &0.956\scriptsize{$\pm$0.020}  & 0.021\scriptsize{$\pm$0.025} &\textbf{0.007\scriptsize{$\pm$0.001}} &0.034\scriptsize{$\pm$0.008}  \\

Spring \& Gravity             &0.798\scriptsize{$\pm$0.011}  &0.219\scriptsize{$\pm$0.310} &\textbf{0.031\scriptsize{$\pm$0.004}}  &0.085\scriptsize{$\pm$0.006} \\
Gravity \& None               &0.096\scriptsize{$\pm$0.004} &0.011\scriptsize{$\pm$0.002} &\textbf{0.008\scriptsize{$\pm$0.001}}  &0.013\scriptsize{$\pm$0.002}  \\
Spring \& Gravity \& None     &0.097\scriptsize{$\pm$0.004}  &0.040\scriptsize{$\pm$0.007} &0.035\scriptsize{$\pm$0.004} & \textbf{0.034\scriptsize{$\pm$0.005}} \\
3 Spring \& None             &0.062\scriptsize{$\pm$0.000}  &0.018\scriptsize{$\pm$0.003} &0.012\scriptsize{$\pm$0.001} & \textbf{0.011\scriptsize{$\pm$0.001}} \\
3 GRavity \& None             
&0.048\scriptsize{$\pm$0.002} 
&0.025\scriptsize{$\pm$0.005} 
&0.021\scriptsize{$\pm$0.001} 
& \textbf{0.014\scriptsize{$\pm$0.001}} \\
100 Spring                &0.113\scriptsize{$\pm$0.002} &0.086\scriptsize{$\pm$0.003} &0.081\scriptsize{$\pm$0.005}  & \textbf{0.044\scriptsize{$\pm$0.001}}  \\
100 Spring \& 100 Gravity            &0.343\scriptsize{$\pm$0.013} &0.291\scriptsize{$\pm$0.001} &0.271\scriptsize{$\pm$0.000} & \textbf{0.080\scriptsize{$\pm$0.002}}  \\
\bottomrule
\end{tabular}
\end{sc}
\end{small}
\end{center}
\vskip -0.1in
\end{table*}

\subsubsection{Relation Reasoning}
We first explored a relational reasoning task. 
The results of estimating the relation type in the physical data using the DSLR and NRI models are shown in Table \ref{table:relation_acc}. Each value represents the accuracy of correctly classifying the types of relations between objects. 
Both the DSLR and NRI models can estimate the relation types with high accuracy, and the superiority of the two models differs depending on the data. 
However, if the number of relations is greater than two (\textit{spring \& gravity \& none, 3 spring \& none, 3 gravity \& none}), the accuracy of our model is significantly superior to that of NRI. 
When the number of relations was incorrectly entered in the NRI, the accuracy was generally low (Table \ref{table:relation_acc} $\bar K = 2K$ row). 

Unlike NRI, where the number of relations must be set in advance, DSLR can infer the number of relations in the system using the silhouette score \cite{silhouette}. 
Table \ref{table:relation_num} shows a silhouette score for each number of relations $K$, i.e., the higher the silhouette score is, the higher the probability that $K$ is proper. 
There are two relations in the system for the first, second, and third datasets in Table \ref{table:relation_num}, three relations for the fourth dataset, and four relations for the last two datasets. 
The DSLR model was able to correctly infer the number of relations for all datasets of the six combinations of relations. 

The relation state inferred by our model can be illustrated in a latent space, as shown in Fig. \ref{Figure:relation_latent_space} after reducing the dimension of the relation state into 2D using a principal component analysis \cite{pca}. 
The points indicating the dimension-reduced relation states are marked with the same color if the relation types are the same. 
The experimental results show that the same relations are located close to each other in the relation latent space, and that different relations are placed far from each other.
Even when the number of relations is large, DSLR can arrange the relation states within the relation latent space to be interpretable, as shown in Fig. \ref{Figure:relation_latent_space} (g) and (h). 
In Fig. \ref{Figure:relation_latent_space} (g), the larger the coefficient of the \textit{spring} is, the more the relation states are located on the left side of the latent space, and the smaller the coefficient is, the more the relation states are on the right side. 
In Fig. \ref{Figure:relation_latent_space} (h), the stronger the \textit{spring} is, the more the relation states are located on the left side of the latent space; the stronger the \textit{gravity} is, the more the relation states are located on the right side; and the weaker the force is, the more likely the relation states gather at the center of the latent space. 

\subsubsection{Future Trajectory Prediction}
The DSLR can predict the future states of the nodes based on the inferred relation states. 
The mean squared errors of the future trajectory predicted by the DSLR and NRI models are listed in Table \ref{table:trajectory_error}. 
For the last two datasets in Table \ref{table:trajectory_error} with a large number of relations, the NRI model was trained by setting the given number of relations $\bar K$ to 8, because in our experiments, NRI with a larger $\bar K$ requires too many computing resources, but does not significantly improve the performance. 

\begin{figure}[t]
\vskip 0.0in
\begin{center}
\centerline{\includegraphics[width=\columnwidth]{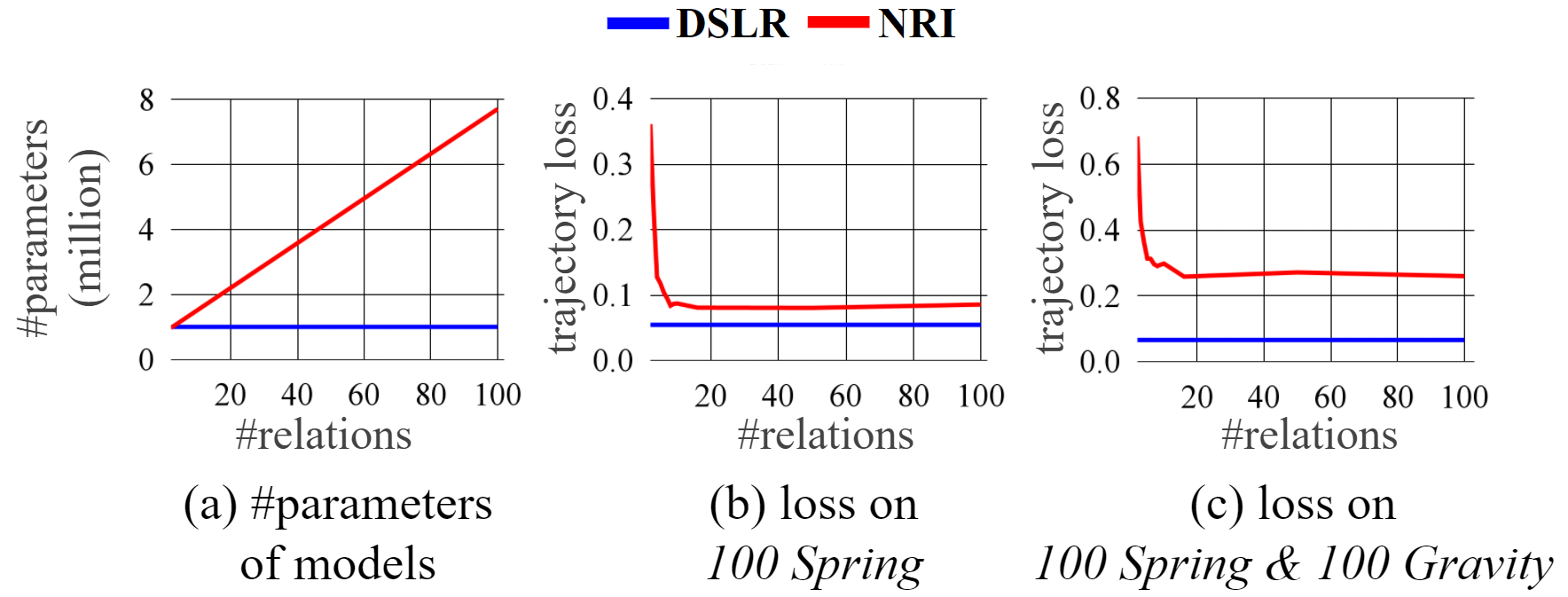}}
\caption{(a) Number of parameters of NRI and DSLR models based on the number of relations given to the model. (b) Prediction errors of future trajectories based on the number of relations given to the model for \textit{100 spring}, and (c) for \textit{100 spring \& 100 gravity}.}
\label{Figure:model_size}
\end{center}
\vskip -0.3in
\end{figure}

\begin{figure}[t]
\vskip 0.0in
\begin{center}
\centerline{\includegraphics[width=\columnwidth]{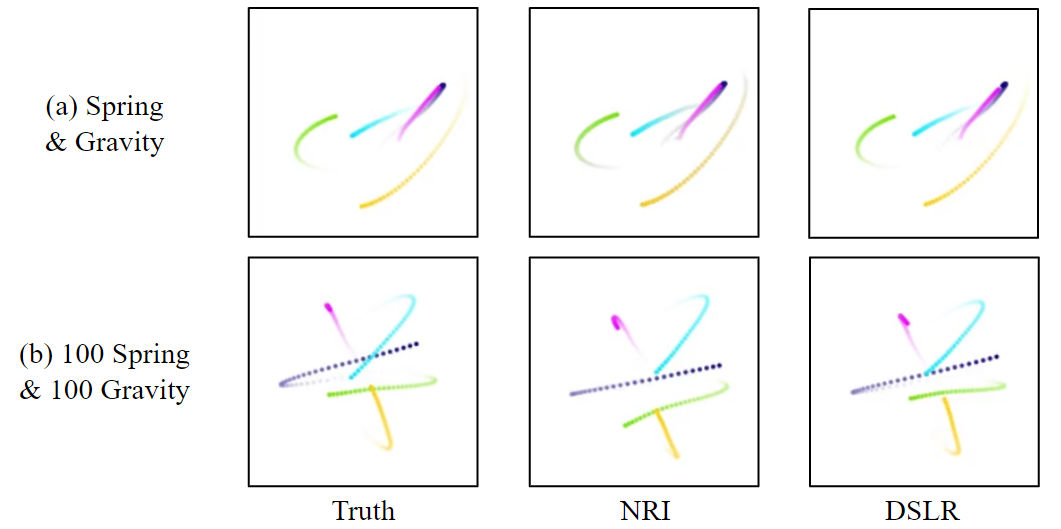}}
\caption{Forecasting the trajectories of physical objects. }
\label{Figure:physics_trajectories}
\end{center}
\vskip -0.3in
\end{figure}

For the first three datasets with a small number of relations, the superiority of the DSLR model and NRI model with correct $\bar K$ differs depending on the combinations of relations. 
Although, the NRI with larger $\bar K$ is less accurate at classifying the relations, NRI can predict the future trajectories better than the other models for the first three datasets (see the row ``$\bar K=2K$'' in Table \ref{table:trajectory_error}). 
If a smaller number of relations are given to the NRI than the number in reality, the error is much larger (see the row ``$\bar K=\lceil K/2 \rceil$'' in Table \ref{table:trajectory_error}). 
However, if the number of relations is larger than two, DSLR outperforms the other models on all datasets. 
In particular, when the relations are almost continuous, as in the last two datasets, DSLR is able to predict the future trajectories much more accurately than the other models. 
Fig. \ref{Figure:model_size} (b) and (c) show the prediction error of the future trajectories of the NRI model for each number of relation types given to the model in the last two datasets. 
We trained the NRI with $\bar K \in \{2,3,4,5,6,7,8,9,10,16,50,100\}$, where the batch size was set to 128 for $\bar K \in [1,16]$, but 32 for $\bar K = 50$, and 16 for $\bar K = 100$ owing to memory limitations. 
As $\bar K$ increases, the future prediction error decreases; however, the number of parameters in the model increases (see Fig. \ref{Figure:model_size} (a)). 
In addition, the errors in the NRI are still greater than the errors in the DSLR, and the number of parameters in the DSLR is lower. 
An example of the results of predicting the future trajectory of an object is shown in Fig. \ref{Figure:physics_trajectories}. 
For data with a small number of relations, both the DSLR and NRI models predicted a future trajectory similar to the truth (see Fig. \ref{Figure:physics_trajectories} (a)), whereas for data with a large number of relations, the DSLR predicted the trajectories more similarly than the NRI (see Fig. \ref{Figure:physics_trajectories} (b)).

\subsubsection{Relation Centrality}
\begin{figure}[t]
\vskip 0.0in
\begin{center}
\centerline{\includegraphics[width=\columnwidth]{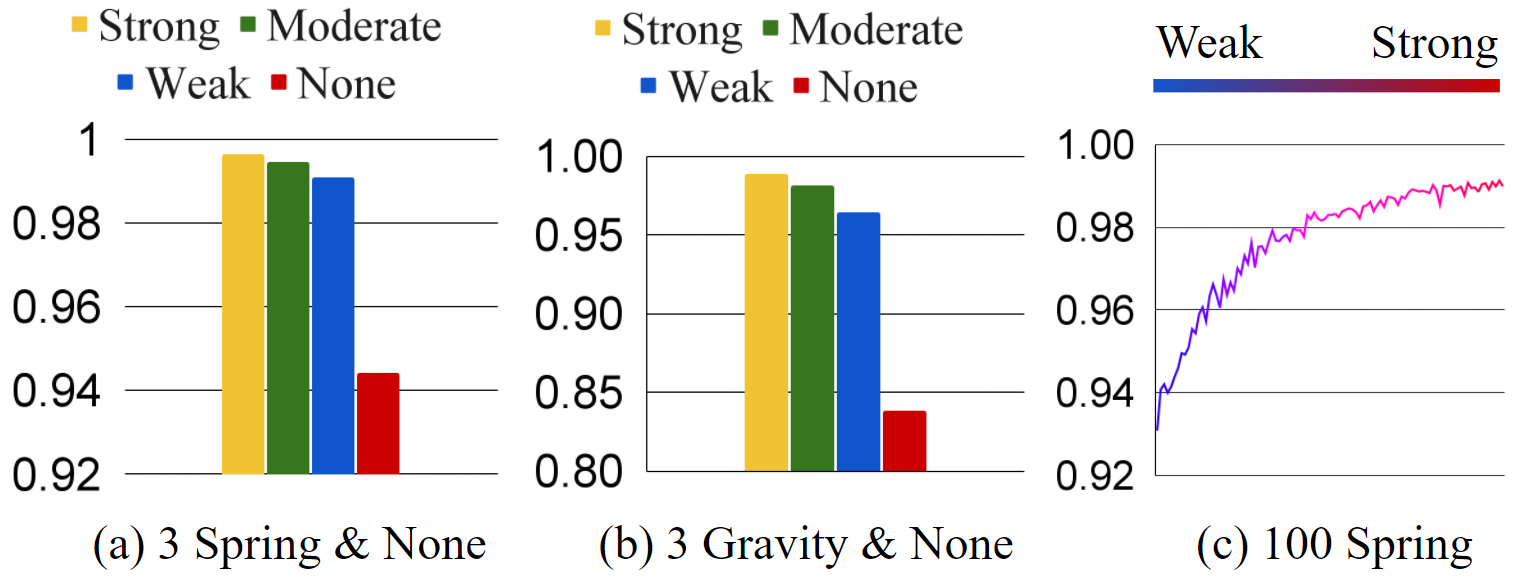}}
\caption{Average relation centrality for each relation type in physically simulated data. Relation centrality for each relation type in (a) \textit{3 spring \& none} and (b) \textit{3 gravity \& none}. (c) Relation centrality from weak to strong spring in \textit{100 spring}.}
\label{Figure:relation_centrality}
\end{center}
\vskip -0.3in
\end{figure}

The DSLR can infer the relation centrality, which is an indicator of the importance of the relation. 
To verify that the relation centrality $c_{ij}$ inferred by our model correctly recognizes the importance of the relation $r_{ij}$, we trained the DSLR by setting $\epsilon$ in \eqref{J} as a random variable, and compared the average relation centrality for each relation type. 
Fig. \ref{Figure:relation_centrality} shows the average relation centrality for each relation type in the \textit{3 spring \& none}, \textit{3 gravity \& none}, and \textit{100 spring} datasets. 
The horizontal axis of the graph indicates the type of relation in the data, and the vertical axis indicates the relation centrality. 
In Fig. \ref{Figure:relation_centrality} (a) and (b), the \textit{none} relationship had the lowest relation centrality, and the stronger the strength of \textit{spring} or \textit{gravity}, the greater the relation centrality. 
Similarly, in Fig. \ref{Figure:relation_centrality} (c), the stronger the strength of \textit{spring} is, the greater the relation centrality. 
Because a relation type with a strong force will have a greater effect on the whole system, it is reasonable that a relation with a strong force will have a higher relation centrality. 
As a result, the experiment shows that the relation centrality inferred by DSLR correctly contains the importance of relations within the system.

\subsection{Motion Capture Data}

We trained the DSLR model using the large motion capture data provided by Carnegie Mellon University \cite{cmuMocap} to infer the relationship between the joints of the human body and predict the future motion of the joints. 
As in previous studies \cite{NRI, dNRI}, we experimented with the walking motion data of the $35$th subject: $11$ trials for training, $4$ trials for validation, and $7$ trials for testing. 
We trained the DSLR model, NRI model \cite{NRI}, and dNRI model \cite{dNRI} with the sparsity prior set to 0.91 as in the previous study \cite{NRI}; and assumed that 91\% of the pairs of joints would have a \textit{none} relation. 
The NRI and dNRI models were trained using four relation types determined experimentally by the authors, one of which was hard-coded for the \textit{none} relation. Whereas the DSLR and NRI models estimated the static relations, dNRI, which is designed to infer the dynamic relations estimated in the experiments, assumed that the relations between human joints may change over time. 
At the inference time, the models observed the system for 49 time-steps to estimate the relation states between joints, and then predicted the future motion of the joints for 50 time-steps. 
Because it is not certain whether the system has a static relation, the relational standard deviation loss was not used, and $m$ in (9) was set to $1$. 
All models were trained for $2000$ epochs. 

\begin{figure}[t]
\vskip 0.0in
\begin{center}
\centerline{\includegraphics[width=1\columnwidth]{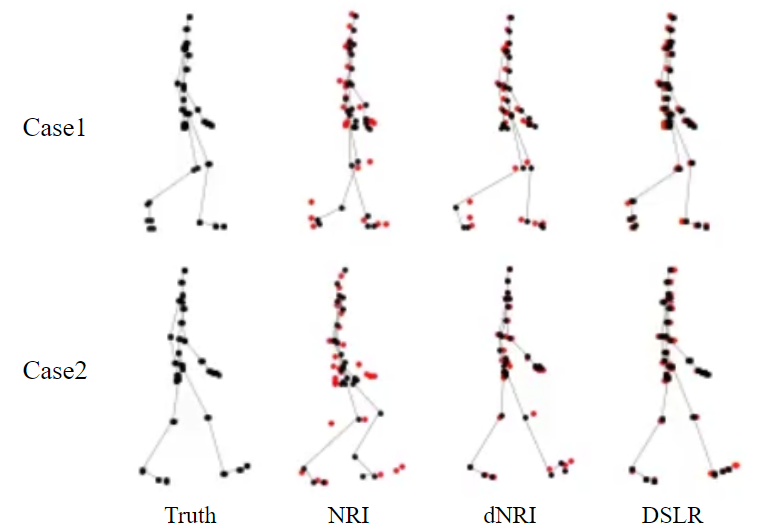}}
\caption{Experimental results with CMU walking mocap data.}
\label{Figure:cmu_trajectories}
\end{center}
\vskip -0.2in
\end{figure}

The first row in Table \ref{table:cmu} represents the errors of the predicted future positions of the joints in the human body as predicted by the DSLR and previous methods. 
The experimental results showed that the DSLR most accurately predicted the future movements of the joints. 
The DSLR can forecast the movement of the joints even better than dNRI, which was designed to infer dynamic relations. 
This is because the DSLR is more suitable for dealing with complex relations in reality because relation states are represented as continuous latent variables, whereas comparative models represent the relations discretely. 
Fig. \ref{Figure:cmu_trajectories} shows two experimental results, where the red dots indicate the true positions of the joints, and the black dots indicate the positions predicted by each model. 
In both cases, the DSLR model predicted the joint movement of the human body would be more similar to the ground truth than the comparative models. 

\begin{figure}[t]
\vskip 0.0in
\begin{center}
\centerline{\includegraphics[width=1\columnwidth]{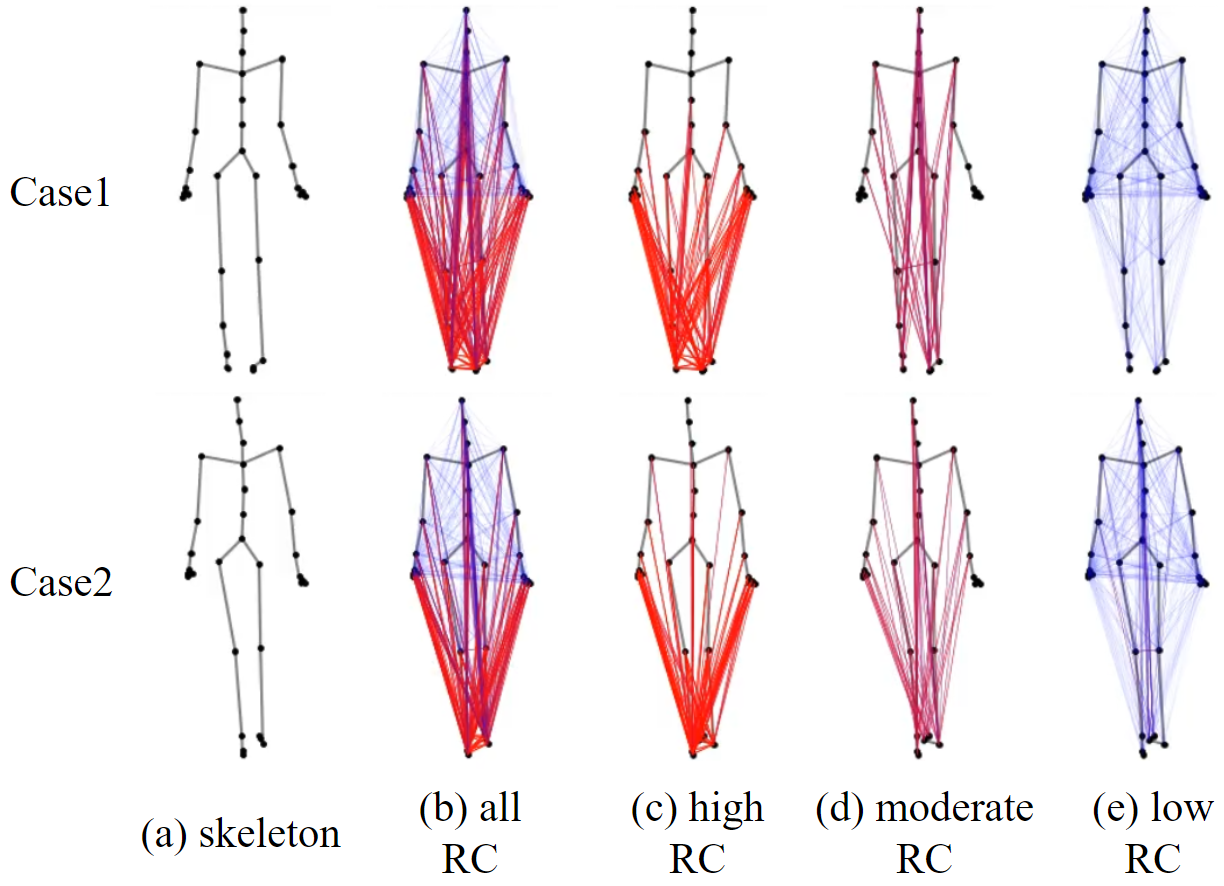}}
\caption{Visualization of relation centrality (RC) in the mocap data inferred using DSLR.  }
\label{Figure:cmu_relation}
\end{center}
\vskip -0.3in
\end{figure}

Fig. \ref{Figure:cmu_relation} shows the visualization of the edge centrality between joints estimated by the DSLR model: the larger the edge centrality is, the thicker the red line, and the smaller the edge centrality is, the thinner the blue line. 
Because the sparsity prior was applied to the model, most of the joints were connected by a weak edge centrality. 
The arms and legs were mainly connected by thick red lines, indicating that the DSLR judged that the relations between arms and legs are the most important during a walking motion (see Fig. \ref{Figure:cmu_relation} (c)).  
In addition, the upper body and legs were connected by a bluish-red line, which means that the DLSR determined that the relations between the upper body and legs are less important than the relations between the arms and legs during a walking motion (see Fig. \ref{Figure:cmu_relation} (d)). 
Finally, other relations are represented as light blue lines, which means that they are the least important relations (see Fig. \ref{Figure:cmu_relation} (e)). 
Although there is no correct answer for the relations between human joints, the interpretation of the DSLR seems to be one of the correct interpretations given that the movements of the arm and legs are most pronounced when a person walks.

Fig. \ref{Figure:relation_latent_space} (i) shows a visualization of the relation states between joints in a two-dimensional relation latent space.
The least important blue relation states are placed on the right side of the latent space, and the most important red relation states are placed vertically across the left side of the latent space. 
Moderately important bluish-red relation states are clustered in the center of the latent space. 

\begin{table}[t]
\caption{Errors of future trajectory prediction for CMU walking mocap data.}
\label{table:cmu}
\vskip 0.1in
\begin{center}
\begin{small}
\begin{sc}
\begin{tabular}{lcccr}
\toprule
Data set & NRI & dNRI & DSLR \\
\midrule
CMU mocap              &0.850\scriptsize{$\pm$0.181}  &0.247\scriptsize{$\pm$0.011}  & \textbf{0.134\scriptsize{$\pm$0.002}}  \\
Basketball              &0.307\scriptsize{$\pm$0.042}   &0.156\scriptsize{$\pm$0.047}  & \textbf{0.079\scriptsize{$\pm$0.001}}  \\
\bottomrule
\end{tabular}
\end{sc}
\end{small}
\end{center}
\vskip -0.2in
\end{table}

\subsection{Basketball Data}
We also experimented with data recording the movements of basketball players \cite{bball}. 
We configured the data with the setup of \cite{dNRI}. 
There are five players in the basketball dataset, and the data consists of 49 time steps, which corresponds to approximately 8 s. 
There are $79456$ training sets, $27690$ validation sets, and $X$ test sets. 
We trained DSLR, NRI \cite{NRI}, and dNRI \cite{dNRI} models and compared the errors of the predicted future trajectories. 
We set the number of relations $\bar K$ to two for NRI and dNRI as determined experimentally by the authors. 
When training the DSLR, $m$ in (9) was set to $1$ as in the motion capture data, and the sparsity prior was not used. 
All models were trained for $1000$ epochs. 

\begin{figure}[t]
\vskip 0.0in
\begin{center}
\centerline{\includegraphics[width=1\columnwidth]{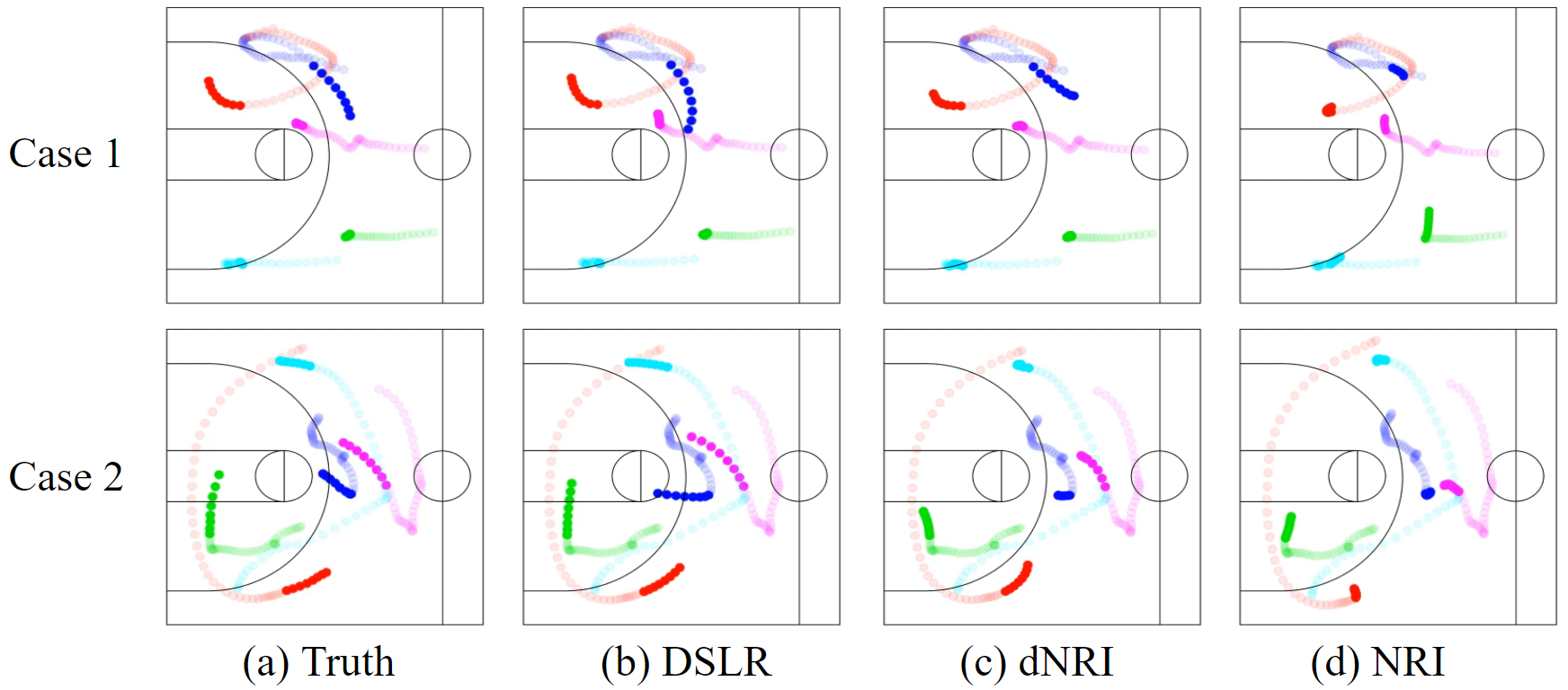}}
\caption{Forecasting the trajectories of basketball players. 
Each player is represented in a different color. 
The transparent points are the trajectories input into the model, and the vivid points are the trajectories predicted by each model.}
\label{Figure:bball_trajectory}
\end{center}
\vskip -0.3in
\end{figure}

The DSLR and comparative models observed the first 40 steps to infer the relationship between players, and then predicted the movement of each player for $9$ steps. 
The second row in Table \ref{table:cmu} represents the average of the prediction errors for the nine steps predicted by each model. 
The numerical results show that the DSLR can predict future trajectories more accurately than the other models. 
Fig. \ref{Figure:bball_trajectory} shows two examples of trajectories predicted by each model in the basketball dataset. 
In the first case, the DSLR model predicted the movement of the red and blue players better than the dNRI, whereas the error of the pink player's movement was larger with the DSLR. 
There was no significant difference between the two models in predicting the movements of the light blue and green players. 
In the second case, DSLR predicted most player movements better than dNRI. 
In both cases, NRI had a larger error than the other models. 
Overall, DSLR was able to predict the movements of players more accurately than the other models, which is congruent with the numerical results. 

\begin{figure}[t]
\vskip 0.0in
\begin{center}
\centerline{\includegraphics[width=0.85\columnwidth]{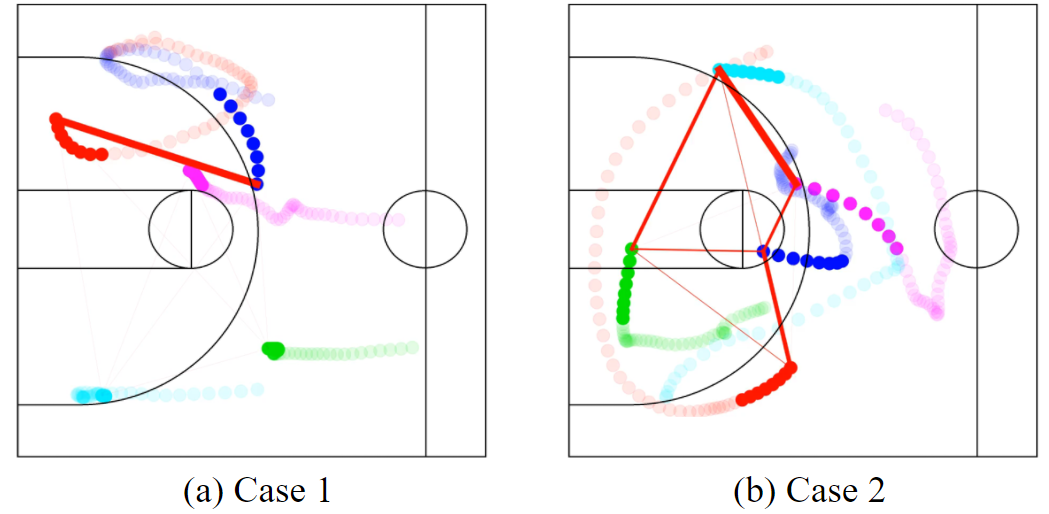}}
\caption{Visualization of relation centrality in the basketball data inferred by DSLR.}
\label{Figure:bball_relation}
\end{center}
\vskip -0.3in
\end{figure}

Fig. \ref{Figure:bball_relation} visualizes the relation centrality inferred by our model in the two cases, where important relations with a relation centrality of $0.985$ or more are represented in red lines; here, the thicker the line is, the higher the relation centrality. 
In the first case, DSLR selected only the relationship between the red and blue players as the most important relationship. 
In the second case, the DSLR method judged the relationship between the light blue and the pink players as being the most important. 
In addition, the DSLR also considered the relationships between the green and light blue players, the pink and the blue players, and the blue and red players as important. 
Because there is no ground-truth of the relationship between players, it is difficult to determine whether the relation centrality inferred by the DSLR is correct; however, it is highly likely to be a meaningful interpretation because the DSLR best predicted the future path based on the inferred relation. 

Fig. \ref{Figure:relation_latent_space} (j) shows a visualization of the relation states between players in a two-dimensional relation latent space. 
The less important blue relation states are placed vertically across the middle of the latent space, and the important red relation states are placed obliquely across the right side of the latent space. 

\subsection{Ablation Study}

\subsubsection{Random Sampling Trick}
We propose the ``random sampling trick'' in Section \ref{subsec:random-sampling-trick} to prevent the model from becoming a compression model when training the DSLR. 
The ``w/o RST'' row in Table \ref{table:ablation} shows the relational reasoning results of the DSLR trained without applying a random sampling trick.
The results of the DSLR trained without a random sampling trick were worse for all datasets than those trained with a random sampling trick. In particular, when the number of relations is large, the performance is significantly improved when a random sampling trick is applied.

\subsubsection{Relation Standard Deviation Loss}
The relation standard deviation loss improves the performance of the relational inference by inducing the relation states to have similar values when they represent the same relation. 
As can be seen in the ``w/o RSDL'' row in Table \ref{table:ablation}, the DSLR trained with the relation standard deviation loss estimated the relations more accurately than the DSLR trained without it.

\begin{table}[t]
\caption{Relational clustering accuracy in \% of DSLR with random sampling trick and relation standard deviation loss (ALL), without random sampling trick (w/o RST), and without relation standard deviation loss (w/o RSDL).}
\label{table:ablation}
\vskip 0.1in
\begin{center}
\begin{small}
\begin{sc}
\begin{tabular}{lccccr}
\toprule
\multirow{2}{*}{Dataset} &\multirow{2}{*}{All} &w/o &w/o \\
 & &RST &RSDL \\
\midrule
Spring \& None              &\textbf{95.33}  &90.18  &93.52   \\
Spring \& Gravity             &\textbf{95.22}  &93.66  &92.05   \\
Gravity \& None               &\textbf{90.55}  &85.43  &86.16             \\
Spring \& Gravity \& None     &\textbf{88.80}  &84.47  &81.26   \\
3 Spring \& None            &\textbf{84.35}  &61.42  &83.20   \\
3 Gravity \& None            &\textbf{74.10}  &65.89  &69.22   \\
\bottomrule
\end{tabular}
\end{sc}
\end{small}
\end{center}
\vskip -0.1in
\end{table}
\section{Conclusion}
\label{sec:conclusion}
In this paper, we propose a DSLR model that can infer the relation between entities that have various relations with each other without supervision and predict the future state of such entities. 
Experimental results using physically simulated data show that the DSLR can analyze the system regardless of the number of relations and infer the number of relations in the system. 
In addition, in experiments using both motion capture data and basketball data, the DSLR can be better applied to real-world data with complex relations between entities.
DSLR can also reflect the sparsity prior or analyze relation centrality. 

DSLR can only model the interacting system in which the relations between entities are static over time. 
However, there may be cases in the real world in which the relationship between entities changes over time.
Future studies can be conducted to extend the DSLR for application to systems with such dynamic relations. 

\appendices

\section{Implementation Details}
In this section, we describe the implementation details of DSLR. 
The source code is available online: https://github.com/dlehgo14/DSLR. 

\subsection{Relation Encoder}
In the relation encoder, the function $\mathcal{F}$ in \eqref{edge_state} which infers the edge state between two nodes is implemented using the gated recurrent unit (GRU) \cite{GRU}. 
We used a 4-layer GRU, and the calculation process in the $i$-th GRU block at time $t$ is as follows:
\begin{equation}
    z_t^i = \sigma({W}_{z}^ix_t + {U}_{z}^is_{t-1}^i + b_z^i),
\end{equation}
\begin{equation}
    r_t^i = \sigma({W}_{r}^ix_t + {U}_{r}^is_{t-1}^i + b_r^i),
\end{equation}
\begin{equation}
    \hat{s}_t^i = tanh(W_h^ix_t + U_h^i(r_t^i*s_{t-1}^i) + b_h^i),
\end{equation}
\begin{equation}
    s_t^i = (1-z_t^i)*s_{t-1}^i + z_t*\hat{s}_t^i,
\end{equation}
where $W$, $U$, and $b$ represent the parameters of the GRU, $x_t$ is the input vector, and $s_t$ is the output vector. 
In addition, $z_t$ is the update gate vector, $r_t$ is the reset gate vector, and $\hat{h}_t$ is the candidate activation vector. 
Moreover, $\sigma$ denotes the sigmoid function, and $*$ denotes the element-wise product. 
The input of the first layer $x_t^1$ is the concatenated vector of the two node states, and the input of the $l$-th layer $x_t^l$ for $l\geq2$ is the output of the previous layer $s_t^{l-1}$. 
The initial edge state $s_{t_E}^i$ is initialized as a zero vector. 
The output vector of the last layer is the edge state. 
There are four dimensions for state of the nodes representing the position and velocity in a $2$-D space. 
In motion capture data, there are six dimensions of the state of the nodes because the joints move in three-dimensional space.  
The number of dimensions of $x_t^i$ is twice the number of dimensions of the node state, as the concatenated vector of the two node states. 
The number of dimensions of the edge state, which is the output of the GRU, was set to $128$. 

The $\mathcal{G, H}$ function in \eqref{relation_state}–\eqref{relation_centrality} that infers the relation state and relation centrality from the edge state is composed of multi-layer perceptrons with skip connections. 
Because $\mathcal{G}$ uses the reparameterization trick, it consists of two parallel networks that output the vectors of the mean and standard deviation of the relation state respectively, which are implemented as a $4$-layer MLP. 
We use ReLU as the activation function and a linear function only for the last layer. 
In addition, $\mathcal{H}$ is also composed of a $4$-layer MLP, with $1$ to $3$ layers shared with the mean network of $\mathcal{G}$, and one last layer attached. 
The number of input dimensions of the MLP is 128, which is the number of dimensions of the edge state, and the number of dimensions of the hidden layers is set to $196$. 
The number of dimensions of the relation state, which is the output of $\mathcal{G}$, is set to $10$, and the relation centrality, which is the output of $\mathcal{H}$, is a $1$-dimensional scalar value. 

\subsection{Relation Decoder}

In the relation decoder, $\mathcal{K}$ in \eqref{influence}, which calculates the influence $f$ exerted by one node on another node, is implemented as a $4$-layer MLP with skip connections such as $\mathcal{G}$. 
The input of $\mathcal{K}$ is a concatenated vector of the states of the two nodes, the relation states between the nodes, and the number of dimensions of the hidden layer is 196. 
The number of dimensions of the output, which is the influence $f$, was set to $100$. 
Next, to aggregate all influences received by the $i$-th node, we sum all influences in an element-wise manner. 

In \eqref{L}, $\mathcal{L}$, which calculates the amount of change in the state of the nodes over time, is implemented in an MLP with skip connections such as $\mathcal{G, K}$. 
The input of the network is the state of the node concatenated with the aggregated influence, and the output is the amount of change in the state of the node, which has the same number of dimensions as the state of the node.



\bibliographystyle{unsrt}
\bibliography{bibList}

\begin{thebibliography}{10}

\bibitem{dynamicgraph}
Frank Harary and Gopal Gupta.
\newblock Dynamic graph models.
\newblock {\em Mathematical and Computer Modelling}, 25(7):79--87, 1997.

\bibitem{NRI}
Thomas Kipf, Ethan Fetaya, Kuan-Chieh Wang, Max Welling, and Richard Zemel.
\newblock Neural relational inference for interacting systems.
\newblock In {\em Int. Conf. Machine Learning}, pages 2688--2697, 2018.

\bibitem{fNRI}
Ezra Webb, Ben Day, Helena Andres-Terre, and Pietro Li{\'o}.
\newblock Factorised neural relational inference for multi-interaction systems.
\newblock {\em arXiv preprint arXiv:1905.08721}, 2019.

\bibitem{SUGAR}
Yaguang Li, Chuizheng Meng, Cyrus Shahabi, and Yan Liu.
\newblock Structure-informed graph auto-encoder for relational inference and
  simulation.
\newblock In {\em Int. Conf. Machine Learning Workshop on Learning and
  Reasoning with Graph-Structured Representations}, 2019.

\bibitem{dNRI}
Colin Graber and Alexander~G Schwing.
\newblock Dynamic neural relational inference.
\newblock In {\em Computer Vision and Pattern Recognition}, pages 8513--8522,
  2020.

\bibitem{gnn1}
Franco Scarselli, Sweah~Liang Yong, Marco Gori, Markus Hagenbuchner, Ah~Chung
  Tsoi, and Marco Maggini.
\newblock Graph neural networks for ranking web pages.
\newblock In {\em The 2005 IEEE/WIC/ACM Int. Conf. Web Intelligence (WI'05)},
  pages 666--672. IEEE, 2005.

\bibitem{gnn2}
F.~{Scarselli}, M.~{Gori}, A.~C. {Tsoi}, M.~{Hagenbuchner}, and
  G.~{Monfardini}.
\newblock The graph neural network model.
\newblock {\em IEEE Trans. Neural Networks}, 20(1):61--80, 2009.

\bibitem{gnn3}
Franco Scarselli, Marco Gori, Ah~Chung Tsoi, Markus Hagenbuchner, and Gabriele
  Monfardini.
\newblock Computational capabilities of graph neural networks.
\newblock {\em IEEE Trans. Neural Networks}, 20(1):81--102, 2008.

\bibitem{gnn4}
Yujia Li, Daniel Tarlow, Marc Brockschmidt, and Richard~S. Zemel.
\newblock Gated graph sequence neural networks.
\newblock In Yoshua Bengio and Yann LeCun, editors, {\em Int. Conf. Learning
  Representations}, 2016.

\bibitem{imageDNN}
Kaiming He, Xiangyu Zhang, Shaoqing Ren, and Jian Sun.
\newblock Deep residual learning for image recognition.
\newblock In {\em Computer Vision and Pattern Recognition}, pages 770--778,
  2016.

\bibitem{langDNN}
Jacob Devlin, Ming-Wei Chang, Kenton Lee, and Kristina Toutanova.
\newblock Bert: Pre-training of deep bidirectional transformers for language
  understanding.
\newblock {\em arXiv preprint arXiv:1810.04805}, 2018.

\bibitem{controlDNN}
David Silver, Aja Huang, Chris~J Maddison, Arthur Guez, Laurent Sifre, George
  Van Den~Driessche, Julian Schrittwieser, Ioannis Antonoglou, Veda
  Panneershelvam, Marc Lanctot, et~al.
\newblock Mastering the game of go with deep neural networks and tree search.
\newblock {\em nature}, 529(7587):484, 2016.

\bibitem{accesscnn}
Shahriar~Rahman Fahim, Dristi Datta, MD. Rafiqul~Islam Sheikh, Sanjay Dey,
  Yeahia Sarker, Subrata~K. Sarker, Faisal~R. Badal, and Sajal~K. Das.
\newblock A visual analytic in deep learning approach to eye movement for
  human-machine interaction based on inertia measurement.
\newblock {\em IEEE Access}, 8:45924--45937, 2020.

\bibitem{relationalInductiveBias}
Peter~W Battaglia, Jessica~B Hamrick, Victor Bapst, Alvaro Sanchez-Gonzalez,
  Vinicius Zambaldi, Mateusz Malinowski, Andrea Tacchetti, David Raposo, Adam
  Santoro, Ryan Faulkner, et~al.
\newblock Relational inductive biases, deep learning, and graph networks.
\newblock {\em arXiv preprint arXiv:1806.01261}, 2018.

\bibitem{commNet}
Sainbayar Sukhbaatar, arthur szlam, and Rob Fergus.
\newblock Learning multiagent communication with backpropagation.
\newblock In D.~Lee, M.~Sugiyama, U.~Luxburg, I.~Guyon, and R.~Garnett,
  editors, {\em Advances in Neural Information Processing Systems}, volume~29,
  pages 2244--2252. Curran Associates, Inc., 2016.

\bibitem{IN}
Peter Battaglia, Razvan Pascanu, Matthew Lai, Danilo~Jimenez Rezende, et~al.
\newblock Interaction networks for learning about objects, relations and
  physics.
\newblock In {\em Advances in Neural Information Processing Systems}, pages
  4502--4510, 2016.

\bibitem{relationalReasoning}
Adam Santoro, David Raposo, David~G Barrett, Mateusz Malinowski, Razvan
  Pascanu, Peter Battaglia, and Timothy Lillicrap.
\newblock A simple neural network module for relational reasoning.
\newblock In {\em Advances in Neural Information Processing Systems},
  volume~30, pages 4967--4976. Curran Associates, Inc., 2017.

\bibitem{vain}
Yedid Hoshen.
\newblock Vain: Attentional multi-agent predictive modeling.
\newblock In {\em Advances in Neural Information Processing Systems}, pages
  2701--2711, 2017.

\bibitem{graphAttention}
Petar Veličković, Guillem Cucurull, Arantxa Casanova, Adriana Romero, Pietro
  Liò, and Yoshua Bengio.
\newblock Graph attention networks.
\newblock In {\em Int. Conf. on Learning Representations}, 2018.

\bibitem{access_self_attention}
Myeongjun Kim, Taehun Kim, and Daijin Kim.
\newblock Spatio-temporal slowfast self-attention network for action
  recognition.
\newblock In {\em IEEE Int. Conf. Image Processing}, pages 2206--2210, 2020.

\bibitem{fewshotGNN}
Victor~Garcia Satorras and Joan~Bruna Estrach.
\newblock Few-shot learning with graph neural networks.
\newblock In {\em Int. Conf. on Learning Representations}, 2018.

\bibitem{rNEM}
Sjoerd van Steenkiste, Michael Chang, Klaus Greff, and J{\"u}rgen Schmidhuber.
\newblock Relational neural expectation maximization: Unsupervised discovery of
  objects and their interactions.
\newblock In {\em Int. Conf. Learning Representations}, 2018.

\bibitem{VIN}
Nicholas Watters, Daniel Zoran, Theophane Weber, Peter Battaglia, Razvan
  Pascanu, and Andrea Tacchetti.
\newblock Visual interaction networks: Learning a physics simulator from video.
\newblock In {\em Advances in Neural Information Processing Systems}, pages
  4539--4547, 2017.

\bibitem{discovering}
Miles Cranmer, Alvaro Sanchez-Gonzalez, Peter Battaglia, Rui Xu, Kyle Cranmer,
  David Spergel, and Shirley Ho.
\newblock Discovering symbolic models from deep learning with inductive biases.
\newblock {\em arXiv preprint arXiv:2006.11287}, 2020.

\bibitem{fluid}
Alvaro Sanchez-Gonzalez, Jonathan Godwin, Tobias Pfaff, Rex Ying, Jure
  Leskovec, and Peter~W Battaglia.
\newblock Learning to simulate complex physics with graph networks.
\newblock {\em arXiv preprint arXiv:2002.09405}, 2020.

\bibitem{visualGrounding}
Yunzhu Li, Toru Lin, Kexin Yi, Daniel Bear, Daniel~L.K. Yamins, Jiajun Wu,
  Joshua~B. Tenenbaum, and Antonio Torralba.
\newblock Visual grounding of learned physical models.
\newblock In {\em Int. Conf. Machine Learning}, 2020.

\bibitem{accessgnn}
Nurul~A Asif, Yeahia Sarker, Ripon~K Chakrabortty, Michael~J Ryan, Md~Hafiz
  Ahamed, Dip~K Saha, Faisal~R Badal, Sajal~K Das, Md~Firoz Ali, Sumaya~I
  Moyeen, et~al.
\newblock Graph neural network: A comprehensive review on non-euclidean space.
\newblock {\em IEEE Access}, 2021.

\bibitem{otherRelation1}
Clive~WJ Granger.
\newblock Investigating causal relations by econometric models and
  cross-spectral methods.
\newblock {\em Econometrica: journal of the Econometric Society}, pages
  424--438, 1969.

\bibitem{otherRelation2}
Scott Linderman and Ryan Adams.
\newblock Discovering latent network structure in point process data.
\newblock In {\em Int. Conf. Machine Learning}, pages 1413--1421, 2014.

\bibitem{otherRelation3}
Scott Linderman, Ryan~P Adams, and Jonathan~W Pillow.
\newblock Bayesian latent structure discovery from multi-neuron recordings.
\newblock {\em Advances in Neural Information Processing Systems},
  29:2002--2010, 2016.

\bibitem{VAE}
Diederik~P Kingma and Max Welling.
\newblock Auto-encoding variational bayes.
\newblock {\em arXiv preprint arXiv:1312.6114}, 2013.

\bibitem{clustering}
A.~K. Jain, M.~N. Murty, and P.~J. Flynn.
\newblock Data clustering: A review.
\newblock {\em ACM Comput. Surv.}, 31(3):264–323, September 1999.

\bibitem{silhouette}
Peter~J Rousseeuw.
\newblock Silhouettes: a graphical aid to the interpretation and validation of
  cluster analysis.
\newblock {\em Journal of computational and applied mathematics}, 20:53--65,
  1987.

\bibitem{adam}
Diederik~P Kingma and Jimmy Ba.
\newblock Adam: A method for stochastic optimization.
\newblock In {\em Int. Conf. Learning Representations}, 2015.

\bibitem{pytorch}
Adam Paszke, Sam Gross, Francisco Massa, Adam Lerer, James Bradbury, Gregory
  Chanan, Trevor Killeen, Zeming Lin, Natalia Gimelshein, Luca Antiga, Alban
  Desmaison, Andreas Kopf, Edward Yang, Zachary DeVito, Martin Raison, Alykhan
  Tejani, Sasank Chilamkurthy, Benoit Steiner, Lu~Fang, Junjie Bai, and Soumith
  Chintala.
\newblock Pytorch: An imperative style, high-performance deep learning library.
\newblock In H.~Wallach, H.~Larochelle, A.~Beygelzimer, F.~d\textquotesingle
  Alch\'{e}-Buc, E.~Fox, and R.~Garnett, editors, {\em Advances in Neural
  Information Processing Systems}, pages 8024--8035. Curran Associates, Inc.,
  2019.

\bibitem{onecyclelr}
Leslie~N Smith and Nicholay Topin.
\newblock Super-convergence: Very fast training of neural networks using large
  learning rates.
\newblock In {\em Artificial Intelligence and Machine Learning for Multi-Domain
  Operations Applications}, volume 11006, page 1100612. International Society
  for Optics and Photonics, 2019.

\bibitem{paig}
Miguel Jaques, Michael Burke, and Timothy Hospedales.
\newblock Physics-as-inverse-graphics: Unsupervised physical parameter
  estimation from video.
\newblock In {\em Int. Conf. Learning Representations}, 2019.

\bibitem{pca}
Karl~Pearson F.R.S.
\newblock Liii. on lines and planes of closest fit to systems of points in
  space.
\newblock {\em The London, Edinburgh, and Dublin Philosophical Magazine and
  Journal of Science}, 2(11):559--572, 1901.

\bibitem{cmuMocap}
CMU.
\newblock Carnegie-mellon motion capture database.
\newblock 2003.

\bibitem{bball}
Yisong Yue, Patrick Lucey, Peter Carr, Alina Bialkowski, and Iain Matthews.
\newblock Learning fine-grained spatial models for dynamic sports play
  prediction.
\newblock In {\em IEEE int. conf. data mining}, pages 670--679. IEEE, 2014.

\bibitem{GRU}
Kyunghyun Cho, Bart Van~Merri{\"e}nboer, Caglar Gulcehre, Dzmitry Bahdanau,
  Fethi Bougares, Holger Schwenk, and Yoshua Bengio.
\newblock Learning phrase representations using rnn encoder-decoder for
  statistical machine translation.
\newblock {\em arXiv preprint arXiv:1406.1078}, 2014.

\end{thebibliography}

\begin{IEEEbiography}[{\includegraphics[width=1in,height=1.25in,clip,keepaspectratio]{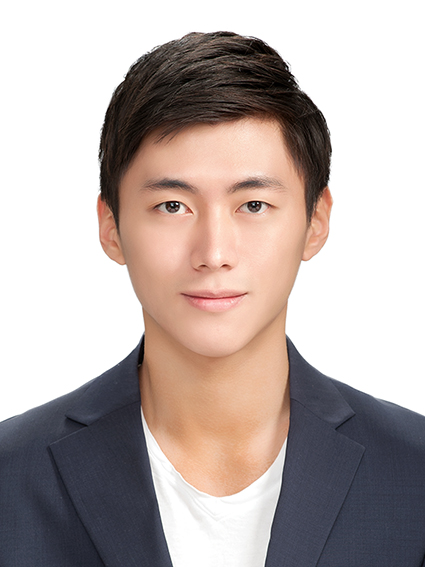}}]{Dohae Lee} was born in Seoul, South Korea in 1995. He received the B.S. degree in food and nutrition, and computer science from the Yonsei University, Seoul, South Korea, in 2019. He is currently a graudate student with the Department of computer science, Yonsei University, Seoul, South Korea. 
\end{IEEEbiography}

\begin{IEEEbiography}[{\includegraphics[width=1in,height=1.25in,clip,keepaspectratio]{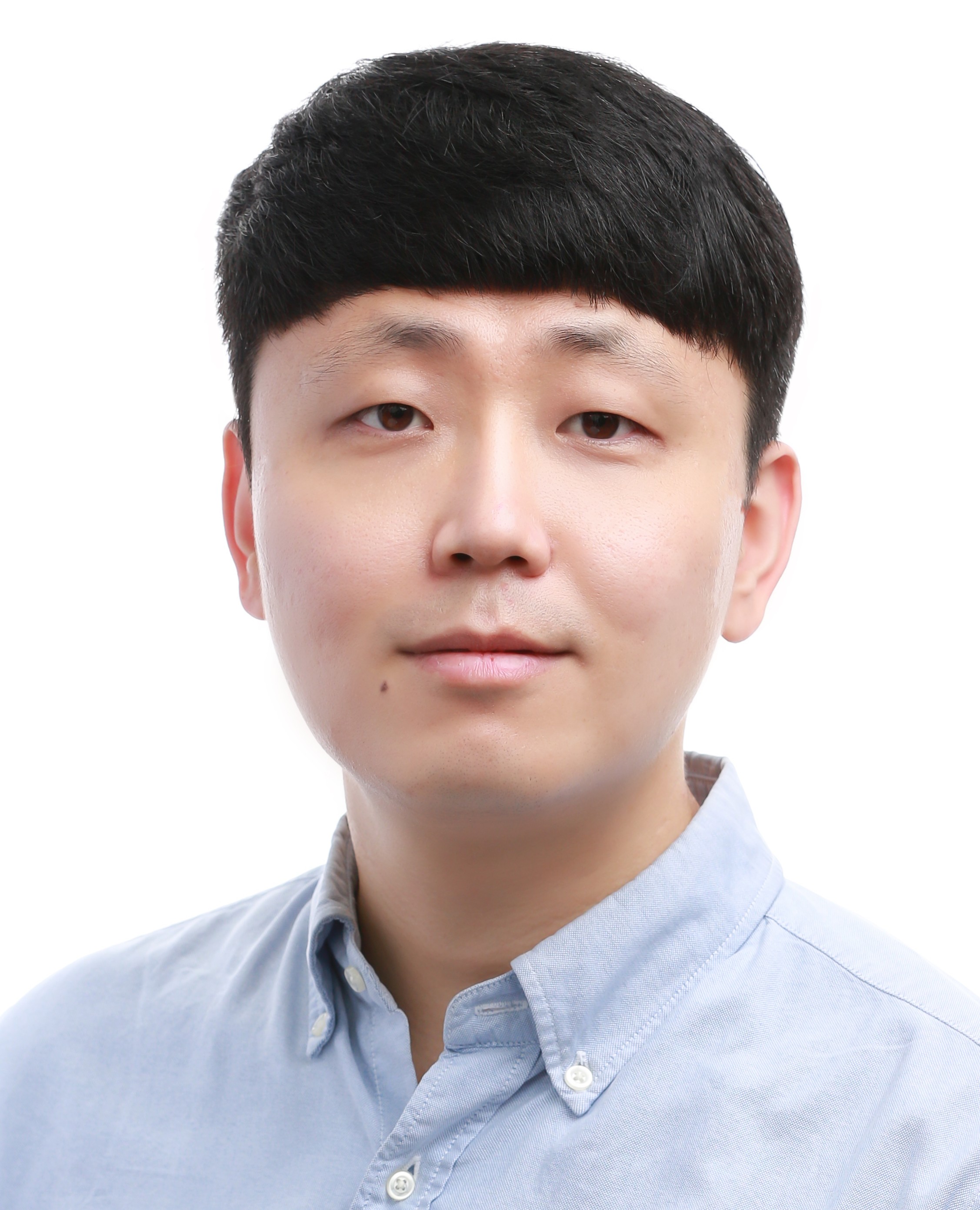}}]{Young Jin Oh} received the B.S. degree in computer engineering from Chungnam National University, Daejeon, 
South Korea, in 2009 and the Ph.D. degree in computer science from Yonsei University, Seoul, South Korea, in 2021. 
He is currently a research engineer in LG Electronics Advanced Robotics Lab.
\end{IEEEbiography}

\begin{IEEEbiography}[{\includegraphics[width=1in,height=1.25in,clip,keepaspectratio]{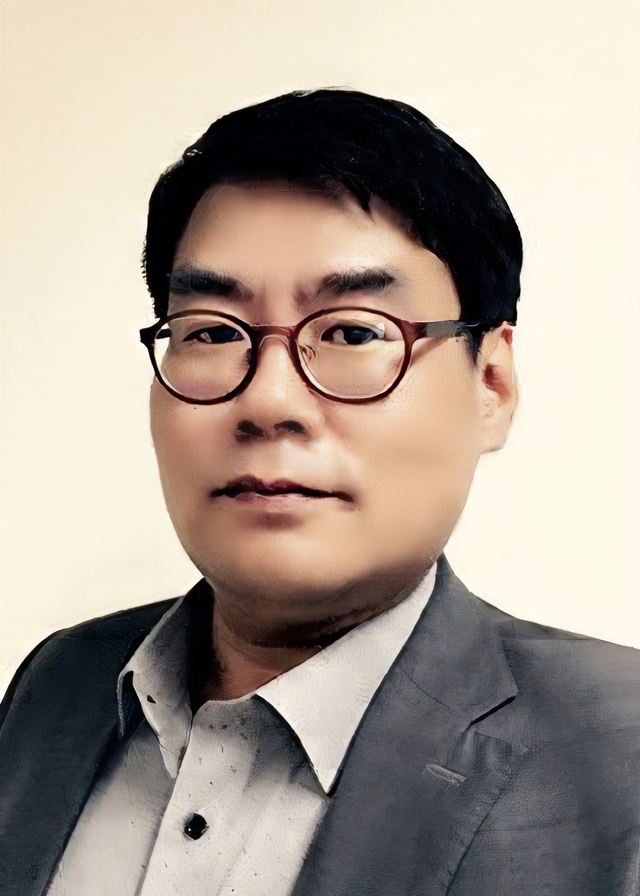}}]{In-Kwon Lee} received the B.S. degree in computer science from Yonsei University, Seoul, South Korea, in 1989, and the M.S. and Ph.D. degrees in computer science and engineering from the Pohang University of Science and Technology, Pohang, South Korea, in
1992 and 1997, respectively. He is a Professor with the Department of Computer Science, Yonsei University. His research interests include computer graphics, HCI and music technology. 
\end{IEEEbiography}

\EOD

\end{document}